\documentclass[conference]{IEEEtran}
\IEEEoverridecommandlockouts
\usepackage{cite}
\usepackage{amsmath,amssymb,amsfonts}
\usepackage{algorithmic}
\usepackage{graphicx}
\usepackage{textcomp}
\usepackage{booktabs}
\usepackage{multirow}
\usepackage{xcolor}
\def\BibTeX{{\rm B\kern-.05em{\sc i\kern-.025em b}\kern-.08em
    T\kern-.1667em\lower.7ex\hbox{E}\kern-.125emX}}
\begin{document}

\title{Pointing-Based Object Recognition}

\author{\IEEEauthorblockN{Lukáš Hajdúch}
\IEEEauthorblockA{
\textit{Comenius University, Bratislava, Slovakia}}
\and
\IEEEauthorblockN{Viktor Kocur}
\IEEEauthorblockA{
\textit{Comenius University, Bratislava, Slovakia}\\
viktor.kocur@fmph.uniba.sk \\ 0000-0001-8752-2685}
}

\maketitle

\begin{abstract}
This paper presents a comprehensive pipeline for recognizing objects targeted by human pointing gestures using RGB images. As human-robot interaction moves toward more intuitive interfaces, the ability to identify targets of non-verbal communication becomes crucial. Our proposed system integrates several existing state-of-the-art methods, including object detection, body pose estimation, monocular depth estimation, and vision-language models. We evaluate the impact of 3D spatial information reconstructed from a single image and the utility of image captioning models in correcting classification errors. Experimental results on a custom dataset show that incorporating depth information significantly improves target identification, especially in complex scenes with overlapping objects. The modularity of the approach allows for deployment in environments where specialized depth sensors are unavailable.
\end{abstract}


\section{Introduction}
Natural and intuitive control forms are increasingly important in human-robot interaction (HRI). In domestic and industrial settings, robots are expected to understand human intent through non-verbal cues. Among these, pointing gestures are perhaps the most fundamental way humans designate objects in shared environments \cite{holladay2014legible}.  


The objective of this work is to design and evaluate a robust pipeline capable of identifying a pointed-at object using a single RGB frame. We propose a pipeline that integrates various state-of-the-art modules to solve the pointing target problem in a practical setting. We specifically investigate how monocular 3D depth reconstruction and textual object descriptions from vision-language models can resolve the spatial ambiguities and classification errors inherent in traditional 2D pipelines.

We evaluate the proposed pipeline on our custom dateset consisting of images of humans pointing at objects placed on a table in front of them. The dataset contains ground truth annotations of pointed-at objects and covers scenes of varying difficulties.

The proposed pipeline includes object detection~\cite{jocher2023yolo,minderer2022owlvit}, computation of the pointing vector, body keypoint localization~\cite{lugaresi2019mediapipe}, depth reconstruction with MoGe~\cite{wang2025moge}, and image captioning using BLIP~\cite{li2022blip}, ViT-GPT2~\cite{singh2022vitgpt2}, and GIT~\cite{wang2022git}. In total, four object detection models and three image captioning models were tested. We found that incorporating depth reconstruction helps reduce the number of false negative cases, and that adding textual descriptions of selected objects increases classification reliability in more complex scenes. The resulting solution is a comprehensive system that combines visual data from the camera, estimation of the spatial arrangement of objects, and auxiliary textual descriptions of selected objects, thereby improving the accuracy of identifying the intended target of the pointing gesture.

\begin{figure*}[t]
    \centering
    \includegraphics[width=\linewidth]{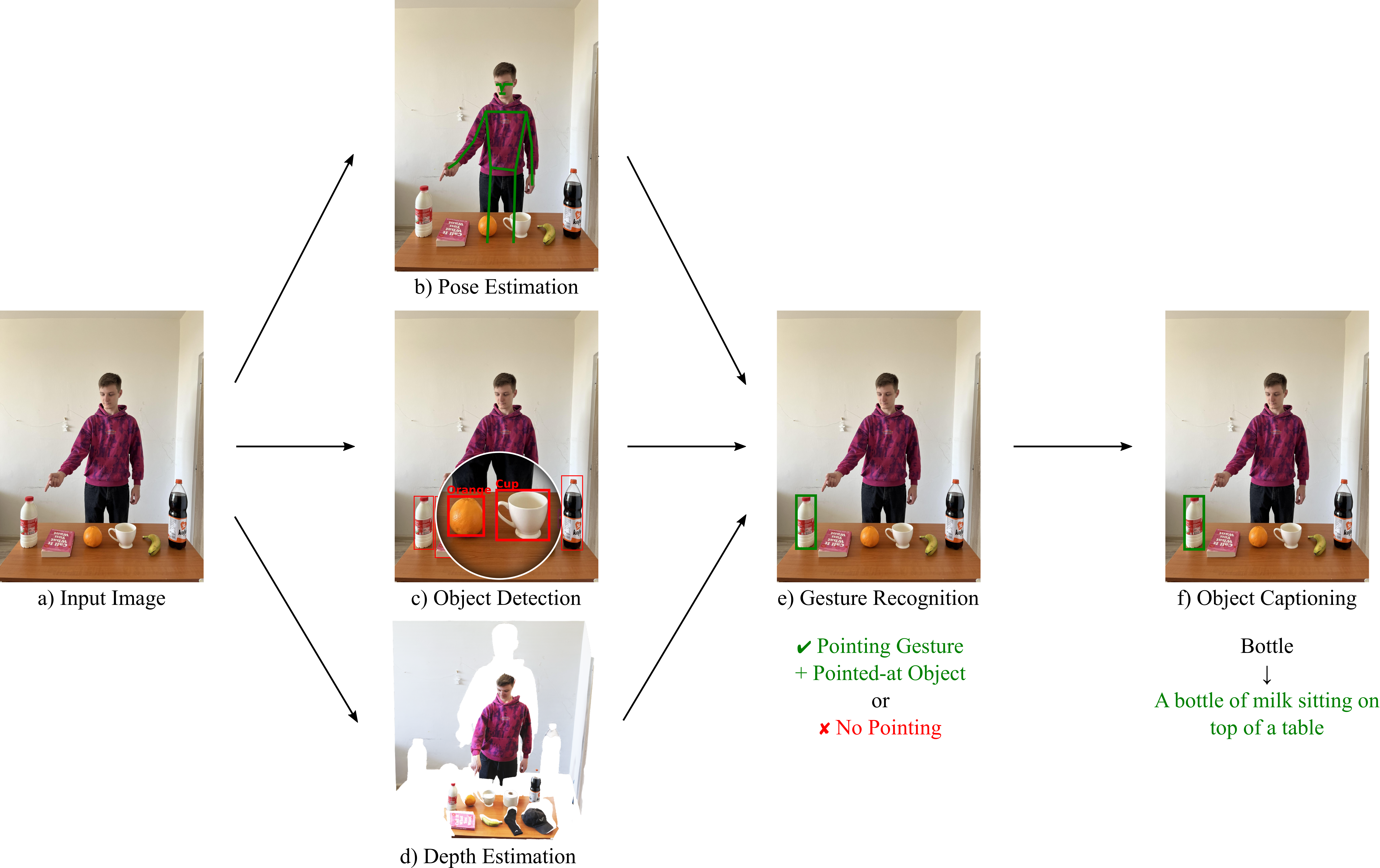}
    \caption{The full pipeline of the proposed system. The input image (a) is fed into three parallel branches: pose estimation (b), object detection (c) and depth estimation (d). The outputs of the three branches get combined to first classify whether pointing occurs and which detected object is pointed at (e). Finally, an image captioning module (f) is used to provide a better description of the object and to potentially fix an incorrect object label assignment.}
    \label{fig:diagram}
\end{figure*}

\section{Related Work}
The field of object recognition has seen a paradigm shift with the advent of Deep Learning \cite{krizhevsky2012imagenet}. Modern architectures like YOLO \cite{redmon2016you,jocher2023yolo} have revolutionized real-time object detection by framing the task as a single regression problem, providing a high balance between inference speed and accuracy. Furthermore, open-vocabulary detectors like OWL-ViT~\cite{minderer2022owlvit} allow for the detection of objects based on natural language queries instead of relying on fixed sets of object categories, making it more versatile in HRI scenarios.

Human pose estimation has similarly progressed, with frameworks like MediaPipe~\cite{lugaresi2019mediapipe} providing efficient solutions for tracking body keypoints. These models allow for the extraction of skeletal data, which is essential for calculating the direction of a pointing gesture. However, 2D pose data alone is often insufficient for accurate pointing direction estimation in complex 3D environments.

Recent advancements in monocular geometry estimation \cite{wang2025moge,piccinelli2024unidepth} have opened new possibilities. Deep neural networks such as MoGe~\cite{wang2025moge} allow for the reconstruction of 3D point clouds from a single 2D image by estimating a depth map and camera intrinsics simultaneously. This allows the system to map 2D image coordinates to full 3D coordinates.

Additionally, the rise of vision-language models~\cite{li2022blip,wang2022git,singh2022vitgpt2} has introduced the concept of image captioning. The captioning models can generate descriptive text for specific regions of an image. Such captioning can be used to enhance the object description for downstream processing. Image captioning can also serve as a verification step by cross-referencing the object detector labels to improve classification reliability.

\section{Proposed Methodology}

The proposed system follows a modular architecture where the input RGB image is processed through three primary parallel branches: object detection, pose estimation, and depth estimation. These outputs are then integrated in a fusion module to identify the target object. The full pipeline is visualized in Fig.~\ref{fig:diagram}. In the text below we describe its individual components.


\subsection{Object Detection Branch}
We employ a variety of object detection models to ensure the system can identify a wide range of household and industrial items. We utilize YOLOv8 and the more recent YOLO11 in various sizes (small, medium, large) provided by Ultralytics~\cite{jocher2023yolo} to test the trade-off between speed and accuracy. For objects not included in the standard COCO dataset~\cite{lin2014microsoft}, we integrate OWL-ViT~\cite{minderer2022owlvit}, which uses a Vision Transformer \cite{dosovitskiy2020vit} backbone to perform open-vocabulary detection based on text prompts. The output of this branch is a set of bounding boxes $B = \{b_1, b_2, ..., b_n\}$ and their associated class labels.

\subsection{Human Pose Recognition}
To accurately identify the user's intent, the system must first extract a reliable representation of the human skeletal structure. We utilize the MediaPipe Pose framework~\cite{lugaresi2019mediapipe}. This model provides 33 3D landmarks, representing the major joints of the human body. We note that the 3D landmarks are provided in normalized coordinates and thus they do not represent the 3D positions of the joints in the scene, but only their relative position to each other.

\subsection{Depth Estimation}

A core component of our pipeline is the use of the MoGe model~\cite{wang2025moge} for monocular depth estimation. For every pixel in the input image, MoGe predicts a depth value, effectively creating a 2D-to-3D mapping. This allows us to transform the 2D centroids of the detected object bounding box $b_i$ into 3D points $\vec{o}_{i}$ in a camera-centered coordinate system. Similarly, we can lift the 2D skeleton landmarks to 3D coordinates. 

\subsection{Object Recognition}

We consider a pointing ray of the form $\vec{p}_w - \vec{p}_s$, which is the difference between the 3D position of the wrist and shoulder. To calculate the pointing score for each object, we also consider the 3D centroid of each detected bounding box $\vec{o}_i$. We then calculate a pointing score $S_i$ for each object $i$:
\begin{equation}
S_i = \frac{\vec{v} \cdot \vec{u}_i}{||\vec{v}|| \cdot ||\vec{u}_i||},
\end{equation}
where $\vec{u}_i = \vec{o}_i - \vec{p}_s$ is the vector from the user's shoulder to the object's 3D centroid. Among the objects we select the one with the highest score as the pointed-at object. This 3D approach significantly reduces the "overlap" problem where a 2D ray might pass through multiple bounding boxes.

In our evaluation we also include a variant of the proposed method where we keep both the pointing direction and the object directions in 2D.

\subsection{Gesture Recognition}

\label{sec:classification}

To identify whether pointing occurs the system relies on the spatial relationship between the 3D positions of shoulder ($\vec{p}_s$), the elbow ($\vec{p}_e$), and the wrist ($\vec{p}_w$). Determining whether the user is actively pointing or simply resting their arm is treated as a binary classification task. We leverage the FLAML library~\cite{wang2021flaml} to optimize a classification model based on the extracted pose features. Two distinct feature representations were evaluated to find the optimal balance between simplicity and precision:
\begin{itemize}
\item \textbf{kpOnly (Keypoints Only):} This set includes the raw $x, y, z$ coordinates of $\vec{p}_s, \vec{p}_e, \vec{p}_w$. These coordinates are normalized relative to the torso size to ensure the model is invariant to the user's distance from the camera.
\item \textbf{full (Extended Feature Set):} Beyond raw coordinates, this set includes engineered geometric features designed to capture the "intent" of the gesture. These include the 3D angle at the elbow joint (to measure arm extension), the Euclidean distance between the wrist and the torso, and the cosine similarity between the forearm and upper arm vectors to quantify the "straightness" of the pointing arm.
\end{itemize}
For both of these cases we also include the best achieved object score among the features.

The resulting classifier allows the system to trigger the object recognition and depth estimation modules only when a high-confidence pointing gesture is detected, thereby reducing computational overhead and minimizing false positives caused by natural body movements.

\subsection{Image Captioning}
\label{sec:captioning}
Standard object detectors often fail on ambiguous items (e.g., distinguishing a white mug from a roll of toilet paper). To address this, we integrated an Image Captioning module. When a pointing gesture is detected, the system crops the target bounding box and passes it to models such as BLIP~\cite{li2022blip}, ViT-GPT2~\cite{singh2022vitgpt2}, or GIT~\cite{wang2022git}. These models generate a textual description of the object. If the caption strongly contradicts the detector's label (e.g., the detector assigns the "cup" label but the caption includes "toilet paper"), the system can flag the result for manual verification or update the label based on the caption's higher semantic understanding.

The captions can be beneficial even when the object detectors assign correct labels as the richer textual representation of the object can potentially be used in further downstream tasks.


\section{Dataset and Experimental Setup}
\begin{figure}
    \centering
    \includegraphics[width=0.3\linewidth]{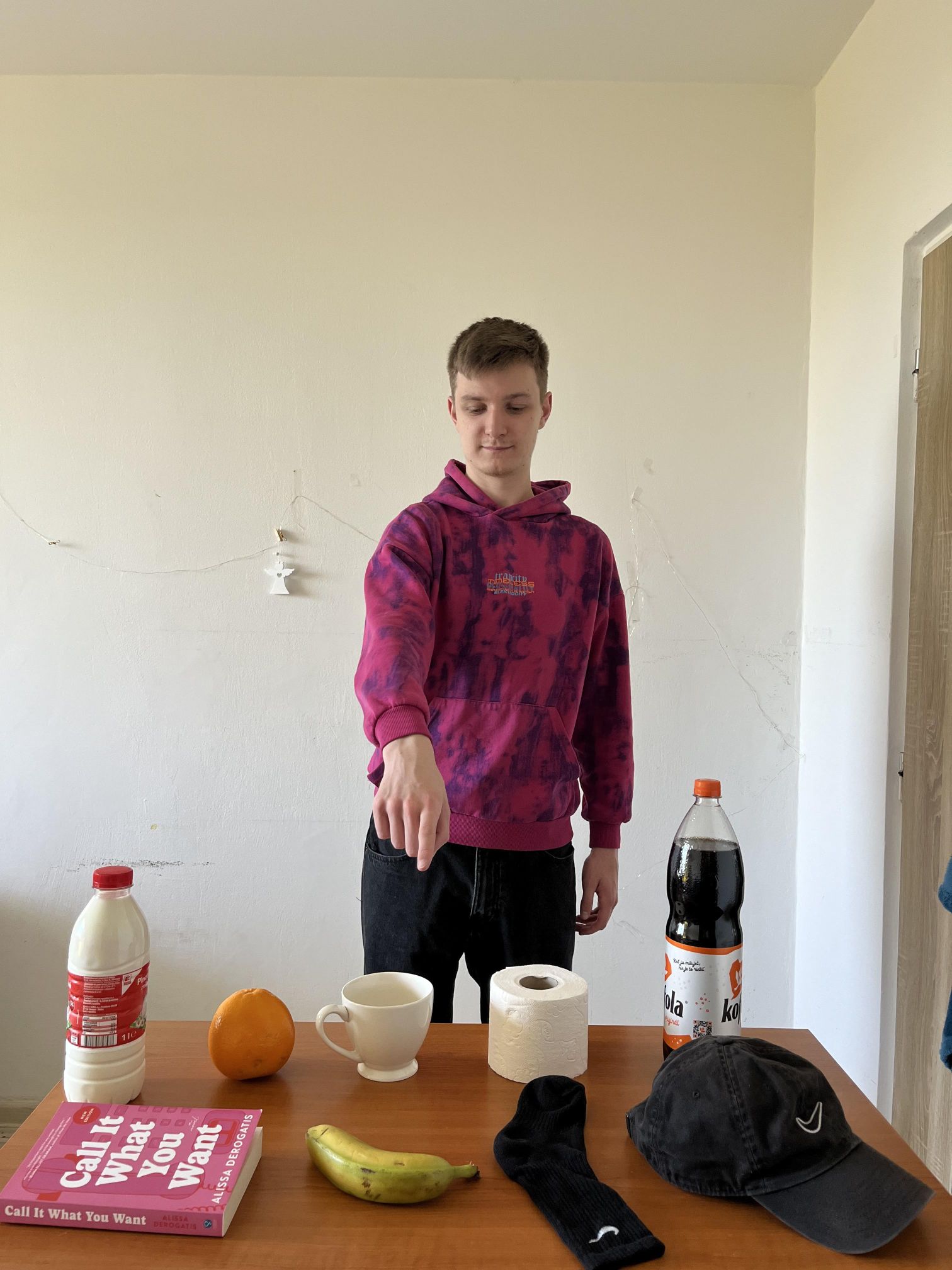}  \hfill \includegraphics[width=0.3\linewidth]{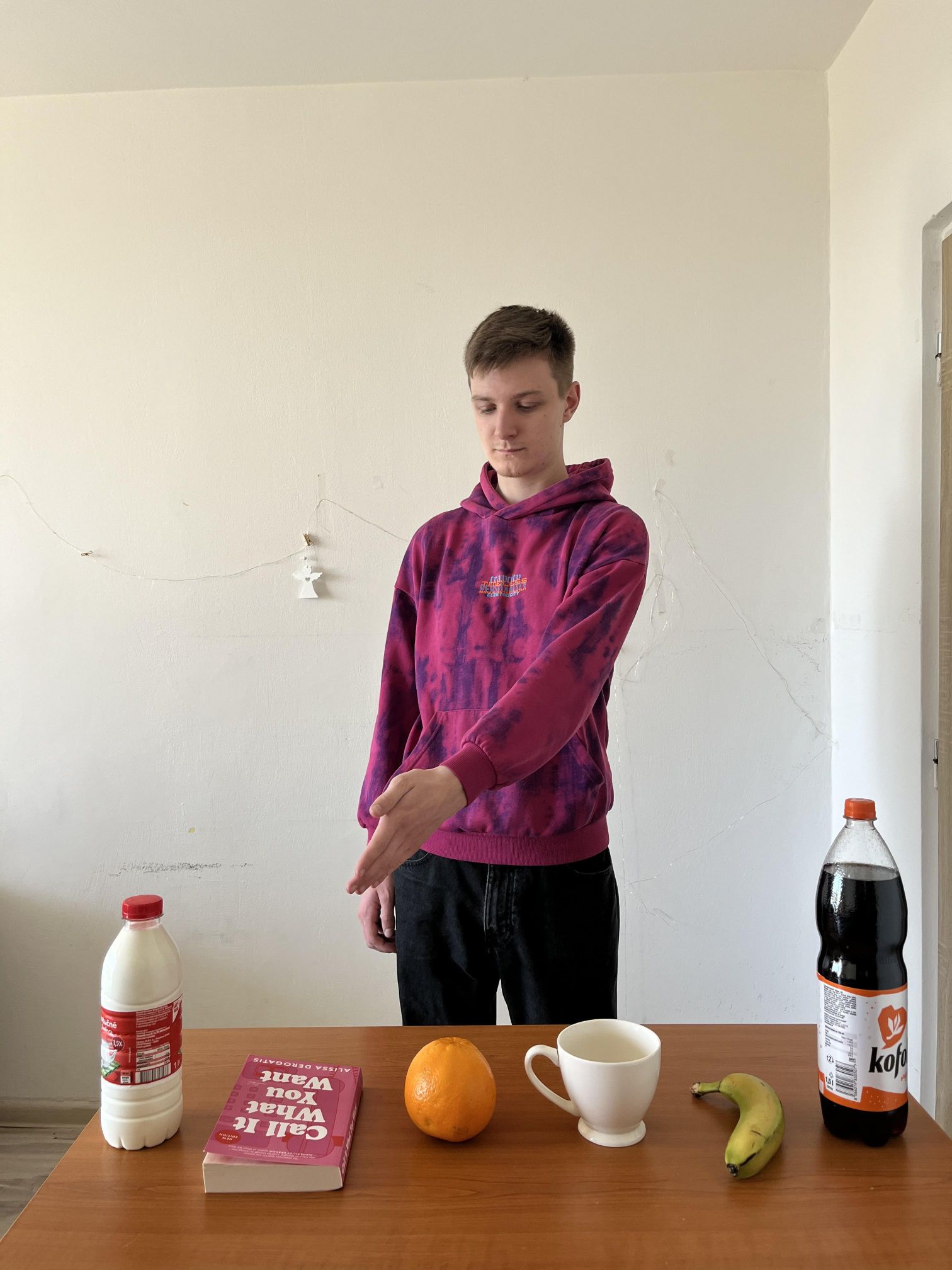}  \hfill     \includegraphics[width=0.3\linewidth]{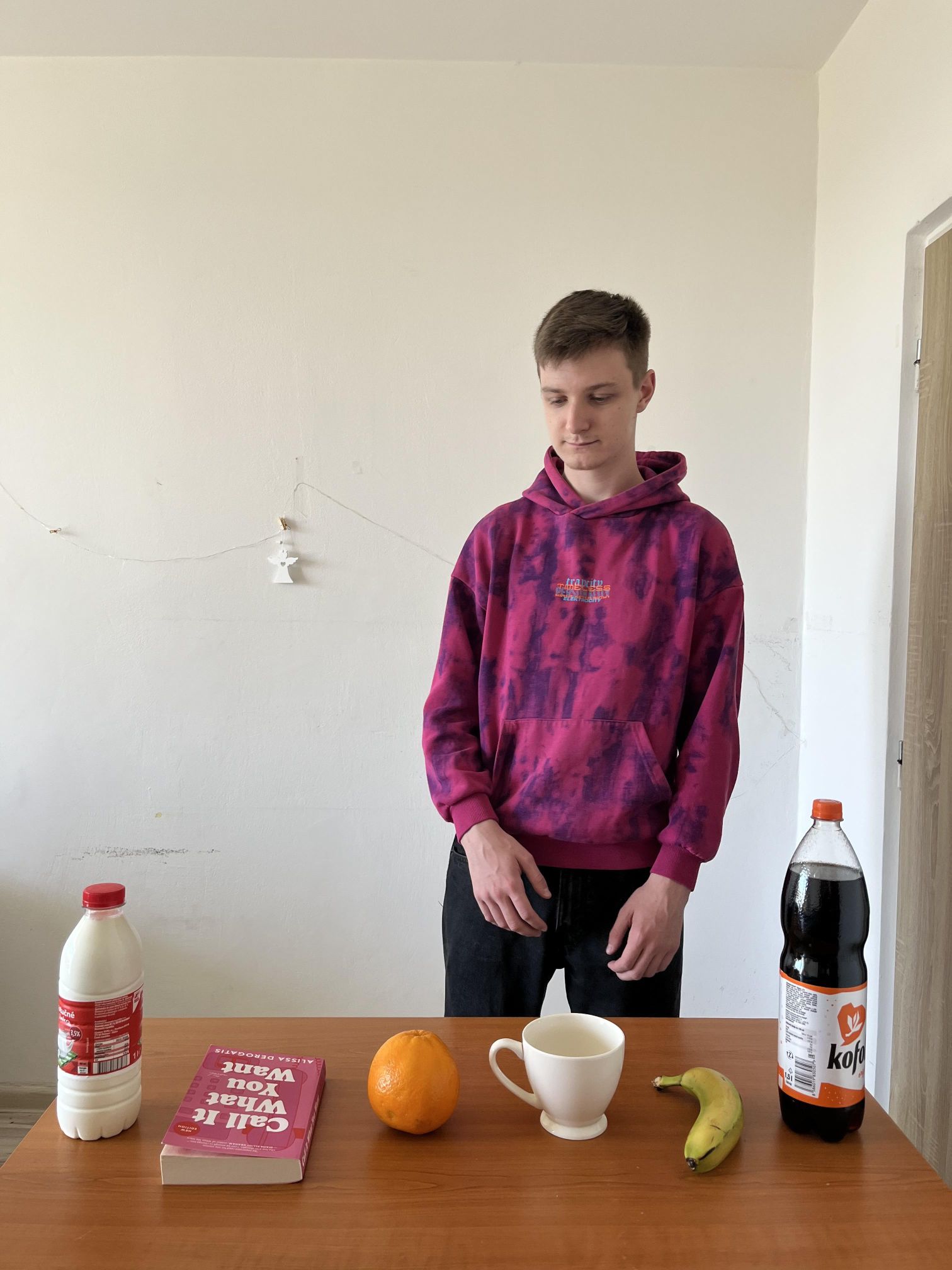}
    \caption{Sample images from our dataset. We use different types of gestures to indicate pointing at objects and neutral poses (no pointing). The dataset captures different levels of complexity in terms of item positions.}
    \label{fig:dataset}
\end{figure}

To evaluate the system, we collected a custom dataset specifically designed to challenge the spatial logic of pointing recognition. The dataset consists of high-resolution RGB frames of users pointing at various objects in a room. For example images see Fig.~\ref{fig:dataset}.

To create the dataset we recruited 5 participants (3 women, 2 men) and instructed them to point at a specific object with one hand, both hands or to assume a neutral pose without pointing at an object. The objects were placed on a table in front of the participant. The data is categorized into two levels of complexity based on object placement:
\begin{enumerate}
    \item \textbf{Easy Cases:} Objects are placed on a single plane (e.g., a table) with significant horizontal spacing (see Fig.~\ref{fig:dataset} right). In these cases, 2D geometry is usually sufficient for identification.
    \item \textbf{Hard Cases:} Objects are arranged in depth (e.g., one bottle behind another see Fig.~\ref{fig:dataset} left) or very close to each other. These scenarios create overlapping bounding boxes in the 2D image, making it impossible to identify the target without depth information.
\end{enumerate}

In total our dataset contains 376 annotated images. We split the data for training and testing. The training set contains 154 images (77 easy cases, 72 hard cases) and the test set contains 222 images (115 easy cases, 107 hard cases). Both sets contain images with participants pointing with either their left hand, right hand, or both.



\section{Results and Discussion}
In this section we present the results for individual components and various configurations of the full proposed pipeline.

\subsection{Gesture Recognition}

\begin{table}[h]
\caption{The classification accuracy results for determining whether pointing occurs in an image (see Sec.~\ref{sec:classification}). The results shown are for combination with YOLO11s~\cite{jocher2023yolo} with ($\checkmark$) and without (\scalebox{1.2}{$\times$}) MoGe depth estimation~\cite{wang2025moge}.}
\centering
\begin{tabular}{lclcccc}
\toprule
\textbf{Case} & \textbf{MoGe} & \textbf{Features} & \textbf{Precision} & \textbf{Recall} & \textbf{F1} & \textbf{Accuracy} \\ \midrule
Easy & $\checkmark$ & kpOnly & 0.53 & 0.86 & 0.66 & 0.77 \\
Easy & $\checkmark$ & full   & \textbf{0.56} & \textbf{0.91} & \textbf{0.69} & \textbf{0.80} \\ \midrule
Hard & $\checkmark$ & kpOnly & \textbf{0.48} & \textbf{0.83} & \textbf{0.61} & \textbf{0.71} \\
Hard & $\checkmark$ & full   & 0.47 & 0.80 & 0.59 & 0.70 \\ \midrule
Easy & \scalebox{1.2}{$\times$} & kpOnly & 0.50 & 0.70 & 0.59 & 0.67 \\
Easy & \scalebox{1.2}{$\times$} & full   & \textbf{0.56} & \textbf{0.67} & \textbf{0.61} & \textbf{0.72} \\ \midrule
Hard & \scalebox{1.2}{$\times$} & kpOnly & 0.35 & 0.55 & 0.42 & 0.54 \\
Hard & \scalebox{1.2}{$\times$} & full   & \textbf{0.46} & \textbf{0.61} & \textbf{0.53} & \textbf{0.65} \\ \bottomrule
\end{tabular}

\label{tab:classification}
\end{table}

We first investigate how well our system is able to detect whether pointing is occurring (see Sec.~\ref{sec:classification}). Table~\ref{tab:classification} shows the results when objects are detected using YOLO11s object detector~\cite{jocher2023yolo}. The results show that using the full feature set is generally better than the kpOnly feature set. Additionally, the results show that incorporating the 3D information from MoGe~\cite{wang2025moge} improves the accuracy significantly. We also evaluated classification accuracy in combination with other object detection models: YOLOv8s, YOLOv8n, YOLO11n~\cite{jocher2023yolo} and OWL-ViT~\cite{minderer2022owlvit} showing very similar results. 

\subsection{Object Recognition}

\begin{table}[h]
\renewcommand{\arraystretch}{1.2} 

\caption{Comparison of object recognition performance across various detection models with ($\checkmark$) and without (\scalebox{1.2}{$\times$}) MoGe depth estimation~\cite{wang2025moge}. Results are categorized into Easy Cases (horizontal separation) and Hard Cases (depth-wise overlap).}

\centering
\begin{tabular}{lccccc}
\hline
\multicolumn{6}{c}{Easy Cases} \\ \hline
\textbf{Model} & \textbf{MoGe} & \textbf{Precision} & \textbf{Recall} & \textbf{F1} & \textbf{Accuracy} \\ \hline
\multirow{2}{*}{YOLOv8s} & $\times$ & 0.42 & 0.49 & 0.45 & 0.63 \\
                         & $\checkmark$ & 0.49 & 0.67 & 0.56 & 0.74 \\ \hline
\multirow{2}{*}{YOLOv8n} & $\times$ & \textbf{0.54} & 0.44 & 0.49 & 0.70 \\
                         & $\checkmark$ & 0.51 & 0.62 & 0.56 & 0.74 \\ \hline
\multirow{2}{*}{YOLO11n} & $\times$ & 0.46 & 0.39 & 0.43 & 0.65 \\
                         & $\checkmark$ & 0.53 & 0.72 & 0.61 & \textbf{0.76} \\ \hline
\multirow{2}{*}{OWL-ViT} & $\times$ & 0.53 & 0.56 & 0.55 & 0.67 \\
                         & $\checkmark$ & 0.51 & \textbf{0.82} & \textbf{0.63} & \textbf{0.76} \\ \hline
\multicolumn{6}{c}{Hard Cases} \\ \hline
\textbf{Model} & \textbf{MoGe} & \textbf{Precision} & \textbf{Recall} & \textbf{F1} & \textbf{Accuracy} \\ \hline
\multirow{2}{*}{YOLOv8s} & $\times$ & 0.32 & 0.35 & 0.33 & 0.55 \\
                         & $\checkmark$ & 0.43 & \textbf{0.75} & \textbf{0.54} & \textbf{0.68} \\ \hline
\multirow{2}{*}{YOLOv8n} & $\times$ & 0.35 & 0.44 & 0.39 & 0.59 \\
                         & $\checkmark$ & 0.41 & 0.70 & 0.52 & \textbf{0.68} \\ \hline
\multirow{2}{*}{YOLO11n} & $\times$ & \textbf{0.44} & 0.43 & 0.44 & 0.64 \\
                         & $\checkmark$ & 0.41 & 0.69 & 0.51 & 0.67 \\ \hline
\multirow{2}{*}{OWL-ViT} & $\times$ & 0.31 & 0.40 & 0.35 & 0.55 \\
                         & $\checkmark$ & 0.39 & 0.74 & 0.51 & 0.66 \\ \hline

\end{tabular}
\label{tab:recognition}
\end{table}

The results for pointed-at object recognition across different detection models and the impact of 3D information are summarized in Table~\ref{tab:recognition}. The experimental data reveals a consistent and significant performance gain when integrating monocular depth estimation via MoGe~\cite{wang2025moge}.

In Easy Cases, where objects are primarily distributed horizontally on a single plane, the transition from 2D to 3D logic provided a noticeable boost in reliability. For instance, using YOLO11n, the accuracy increased from $0.65$ to $0.76$ when MoGe was enabled. Similarly, OWL-ViT achieved its highest F1-score of $0.63$ and a peak recall of $0.82$ in the 3D configuration, suggesting that even in simple scenes, including depth information is beneficial.

The advantage of the 3D approach is most pronounced in Hard Cases, characterized by depth-wise overlapping objects. In these scenarios, 2D-only methods struggle significantly because a single pointing vector often pierces multiple overlapping bounding boxes. The 2D methods saw accuracy drops to as low as $0.55$, while the 3D-integrated models maintained more stable performance, with YOLOv8s and YOLOv8n both reaching $0.68$ accuracy. Moreover, for YOLOv8s, the recall more than doubled from $0.35$ to $0.75$ when 3D information was used.

Among the tested detectors, YOLO11n and OWL-ViT paired with MoGe performed best. OWL-ViT demonstrated superior recall ($0.82$ in Easy, $0.74$ in Hard), which is particularly beneficial for HRI applications where missing a user's intent is often more detrimental than a slight decrease in precision. The results confirm that regardless of the underlying detection architecture, the reconstruction of 3D spatial relationships is the primary driver for resolving ambiguity in monocular pointing gesture recognition.

\subsection{Performance of Image Captioning}

\begin{table}[h]
\centering
\renewcommand{\arraystretch}{1.2} 

\caption{Pointed-at object recognition results for YOLO11s~\cite{jocher2023yolo} object detector combined with different image captioning methods used for correcting badly classified objects (see Sec.~\ref{sec:captioning}).}

\begin{tabular}{llcccc}
\hline
\textbf{Case} & \textbf{Captioning} & \textbf{Precision} & \textbf{Recall} & \textbf{F1} & \textbf{Accuracy} \\ \hline
\multirow{4}{*}{Easy} & \scalebox{1.2}{$\times$} & 0.56 & 0.91 & 0.69 & 0.80 \\
 & BLIP & \textbf{0.58} & \textbf{0.92} & \textbf{0.71} & \textbf{0.81} \\ 
 & ViT-GPT2 & 0.56 & 0.91 & 0.69 & 0.80 \\
 & GIT & \textbf{0.58} & \textbf{0.92} & \textbf{0.71} & \textbf{0.81} \\ \hline
\multirow{4}{*}{Hard} & \scalebox{1.2}{$\times$} & 0.47 & 0.80 & 0.59 & 0.70 \\
 & BLIP & \textbf{0.51} & \textbf{0.81} & \textbf{0.62} & \textbf{0.71} \\
 & ViT-GPT2 & 0.48 & 0.80 & 0.60 & \textbf{0.71} \\
 & GIT & 0.47 & 0.80 & 0.59 & 0.70 \\ \hline
\end{tabular}
\label{tab:captioning}
\end{table}

In several instances, both OWL-ViT~\cite{minderer2022owlvit} and YOLO~\cite{jocher2023yolo} misclassified objects due to lighting or unusual angles. For example, a pair of folded socks was occasionally identified as "sports ball" by the detector. However, the GIT model~\cite{wang2022git} correctly generated the caption "a pair of socks on a table," which allowed the system to correct the classification. The results for our pipeline with various image captioning models in combination with YOLO11s~\cite{jocher2023yolo} are shown in Table~\ref{tab:captioning}. The BLIP image captioning network~\cite{li2022blip} performed the best in terms of improved accuracy of the method.

While the captioning models are computationally more expensive and slower than the detectors, they only need to be triggered once a pointing gesture is confirmed, making them viable for near-real-time applications.

\subsection{Limitations}
Despite the improvements, the system faces challenges with extreme distances. As the distance between the user and the object increases, small errors in the estimated pointing vector result in large displacements at the target, sometimes causing the system to miss the object entirely. Additionally, the monocular depth estimation, while impressive, still lacks the millimeter precision of hardware-based depth sensors.


\section{Conclusion}
This paper presents an integrated pipeline for pointing-based object recognition using only monocular RGB data. By combining modern object detectors, pose estimation, monocular depth estimation, and vision-language models, we created a system that can resolve spatial ambiguities in 3D space without the need for specialized hardware. 

The results demonstrate that while 2D methods are sufficient for simple layouts, 3D reconstruction is essential for handling complex, real-world environments with overlapping objects. The use of image captioning as a secondary classification check further enhances the reliability of the system. Future work may focus on integrating temporal information from video streams to smooth out jitter in the pointing vector and exploring the use of gaze estimation to further refine the user's intent.

\section*{Acknowledgment}

This work was funded by the EU NextGenerationEU through the Recovery and Resilience Plan for Slovakia under Project No.09I02-03-V01-00029 and under Project No. 09I01-03-V04-00048.

\bibliographystyle{IEEEtran}
\bibliography{lit}

@article{krizhevsky2012imagenet,
  title={Imagenet classification with deep convolutional neural networks},
  author={Krizhevsky, Alex and Sutskever, Ilya and Hinton, Geoffrey E},
  journal={Advances in neural information processing systems},
  volume={25},
  year={2012}
}

@inproceedings{redmon2016you,
  title={You only look once: Unified, real-time object detection},
  author={Redmon, Joseph and Divvala, Santosh and Girshick, Ross and Farhadi, Ali},
  booktitle={Proceedings of the IEEE conference on computer vision and pattern recognition},
  pages={779--788},
  year={2016}
}

@software{jocher2023yolo,
  author = {Jocher, Glenn and Qiu, Jing and Chaurasia, Ayush},
  license = {AGPL-3.0},
  month = jan,
  title = {{Ultralytics YOLO}},
  url = {https://github.com/ultralytics/ultralytics},
  version = {11},
  year = {2025}
}

@inproceedings{minderer2022owlvit,
  title={Simple open-vocabulary object detection},
  author={Minderer, Matthias and Gritsenko, Alexey and Stone, Austin and Neumann, Maxim and Weissenborn, Dirk and Dosovitskiy, Alexey and Mahendran, Aravindh and Arnab, Anurag and Dehghani, Mostafa and Shen, Zhuoran and others},
  booktitle={European conference on computer vision},
  pages={728--755},
  year={2022},
  organization={Springer}
}

@article{lugaresi2019mediapipe,
  title={Mediapipe: A framework for building perception pipelines},
  author={Lugaresi, Camillo and Tang, Jiuqiang and Nash, Hadon and McClanahan, Chris and Uboweja, Esha and Hays, Michael and Zhang, Fan and Chang, Chuo-Ling and Yong, Ming Guang and Lee, Juhyun and others},
  journal={arXiv preprint arXiv:1906.08172},
  year={2019}
}

@inproceedings{wang2025moge,
  title={Moge: Unlocking accurate monocular geometry estimation for open-domain images with optimal training supervision},
  author={Wang, Ruicheng and Xu, Sicheng and Dai, Cassie and Xiang, Jianfeng and Deng, Yu and Tong, Xin and Yang, Jiaolong},
  booktitle={Proceedings of the IEEE/CVF Conference on Computer Vision and Pattern Recognition},
  pages={5261--5271},
  year={2025}
}

@inproceedings{piccinelli2024unidepth,
  title={Unidepth: Universal monocular metric depth estimation},
  author={Piccinelli, Luigi and Yang, Yung-Hsu and Sakaridis, Christos and Segu, Mattia and Li, Siyuan and Van Gool, Luc and Yu, Fisher},
  booktitle={Proceedings of the IEEE/CVF Conference on Computer Vision and Pattern Recognition},
  pages={10106--10116},
  year={2024}
}

@inproceedings{li2022blip,
  title={Blip: Bootstrapping language-image pre-training for unified vision-language understanding and generation},
  author={Li, Junnan and Li, Dongxu and Xiong, Caiming and Hoi, Steven},
  booktitle={International conference on machine learning},
  pages={12888--12900},
  year={2022},
  organization={PMLR}
}

@article{wang2022git,
  title={Git: A generative image-to-text transformer for vision and language},
  author={Wang, Jianfeng and Yang, Zhengyuan and Hu, Xiaowei and Li, Linjie and Lin, Kevin and Gan, Zhe and Liu, Zicheng and Liu, Ce and Wang, Lijuan},
  journal={arXiv preprint arXiv:2205.14100},
  year={2022}
}

@misc  {singh2022vitgpt2,
    author       = {Singh, Ankur},
    title        = {vit-gpt2-image-captioning},
    year         = 2022,
    url          = { https://huggingface.co/nlpconnect/vit-gpt2-image-captioning },
    doi          = { 10.57967/hf/0222 },
    publisher    = { Hugging Face }
}

@article{dosovitskiy2020vit,
  title={An image is worth 16x16 words: Transformers for image recognition at scale},
  author={Dosovitskiy, Alexey and Beyer, Lucas and Kolesnikov, Alexander and Weissenborn, Dirk and Zhai, Xiaohua and Unterthiner, Thomas and Dehghani, Mostafa and Minderer, Matthias and Heigold, Georg and Gelly, Sylvain and others},
  journal={arXiv preprint arXiv:2010.11929},
  year={2020}
}

@article{wang2021flaml,
  title={Flaml: A fast and lightweight automl library},
  author={Wang, Chi and Wu, Qingyun and Weimer, Markus and Zhu, Erkang},
  journal={Proceedings of machine learning and systems},
  volume={3},
  pages={434--447},
  year={2021}
}

@inproceedings{holladay2014legible,
  title={Legible robot pointing},
  author={Holladay, Rachel M and Dragan, Anca D and Srinivasa, Siddhartha S},
  booktitle={The 23rd IEEE International Symposium on robot and human interactive communication},
  pages={217--223},
  year={2014},
  organization={IEEE}
}

@inproceedings{lin2014microsoft,
  title={Microsoft coco: Common objects in context},
  author={Lin, Tsung-Yi and Maire, Michael and Belongie, Serge and Hays, James and Perona, Pietro and Ramanan, Deva and Doll{\'a}r, Piotr and Zitnick, C Lawrence},
  booktitle={European conference on computer vision},
  pages={740--755},
  year={2014},
  organization={Springer}
}

\end{document}